\documentclass{article}

\PassOptionsToPackage{numbers, compress}{natbib}

 \usepackage[main, final]{neurips_2026}

\usepackage[utf8]{inputenc} 
\usepackage[T1]{fontenc}    
\usepackage{hyperref}       
\usepackage{url}            
\usepackage{booktabs}       
\usepackage{amsfonts}       
\usepackage{nicefrac}       
\usepackage{microtype}      
\usepackage{xcolor}         
\usepackage{amsmath}
\usepackage{multirow}
\usepackage{adjustbox}
\usepackage[table]{xcolor} 
\usepackage{graphicx} 
\usepackage{wrapfig}

\usepackage{subcaption}

\addtocontents{toc}{\protect\setcounter{tocdepth}{0}}

\title{GeoRoPE: Ground-Aware Rotary Adaptation for Remote Sensing Foundation Models}

\author{
Yu Luo$^{1}$ \quad
Kun Hu$^{2}$ \quad
Mengwei He$^{1}$ \quad
Xiaogang Zhu$^{3}$ \quad
Zeng Shan$^{4}$ \quad
Allen Benter$^{5}$ \quad \\
\textbf{Wei Xiang}$^{6}$ \quad 
\textbf{Patrick Filippi}$^{1}$ \quad
\textbf{Thomas Francis Bishop}$^{1}$  \quad 
\textbf{Zhiyong Wang}$^{1}$ \\[4pt]
$^{1}$The University of Sydney \quad 
$^{2}$Edith Cowan University \quad \\
$^{3}$Adelaide University \quad
$^{4}$ Wuhan Polytechnic University \quad \\ 
$^{5}$ Climate, Orange Agricultural Institute\quad 
$^{6}$La Trobe University\\
}

\begin{document}

\maketitle

\begin{abstract}
Remote-sensing foundation models (RSFMs) benefit from pretraining on imagery from multiple sensors and ground sampling distances (GSDs), but such exposure alone does not resolve scale mismatch during downstream adaptation.
A fixed token-grid offset can correspond to different ground distances across sensors, making grid-based positional priors physically inconsistent. 
Meanwhile, heterogeneous spatial granularity means that compact urban regions and homogeneous landscapes may require different positional sensitivities even under the same GSD.
Therefore, we propose {GeoRoPE}, a ground-aware, RoPE-compatible, and parameter-efficient spatial adaptation method for RSFMs. 
GeoRoPE recalibrates token-level positional interactions from two complementary aspects. 
First, \textit{Geo-Coordinate Calibration (GCC)} rescales raw token-grid offsets according to the ground distance represented by one token-grid step, producing geo-calibrated relative coordinates across GSDs. 
Second, \textit{Geo-Frequency Calibration (GFC)} adjusts the native RoPE frequency with a relation-specific factor, enabling position sensitive adaptation to scene-dependent spatial granularity. 
GeoRoPE is injected into pretrained RSFMs through a lightweight adapter, preserving the frozen spatial prior while adding geo-aware positional corrections. 
Experiments across multiple RSFMs, sensors, resolutions, and downstream tasks demonstrate that GeoRoPE improves cross-resolution robustness and scale-sensitive representation learning. 
\footnote{Our code will be publicly available. }
\end{abstract}

\section{Introduction}
\label{sec:intro}

Advances in Earth observation~(EO) have produced imagery across a wide range of ground sample distances~(GSDs), from sub-meter aerial imagery to tens-of-meters satellite imagery.
This diversity has motivated the rapid development of remote-sensing foundation models (RSFMs)~\cite{astruc2024omnisat, do2025robsense, astruc2025anysat}, which are pretrained on cross-sensor and multi-resolution imagery to support transferable representations for downstream tasks~\cite{bonafilia2020sen1floods11, daudt2018urban, benson2024multi}.
Despite this progress, adapting RSFMs across sensor resolutions remains challenging, especially for dense prediction tasks such as semantic segmentation. 
A key reason is that spatial attention in pretrained models often operates on token-grid coordinates, while the ground distance represented by one token-grid step changes with the input GSD. 
As a result, the same token-grid offset may correspond to different physical distances across images, and spatial priors learned at one resolution may be reused at mismatched ground extents under another.

\begin{wrapfigure}{r}{0.55\textwidth} 
    \centering
    \includegraphics[width=0.95\linewidth]{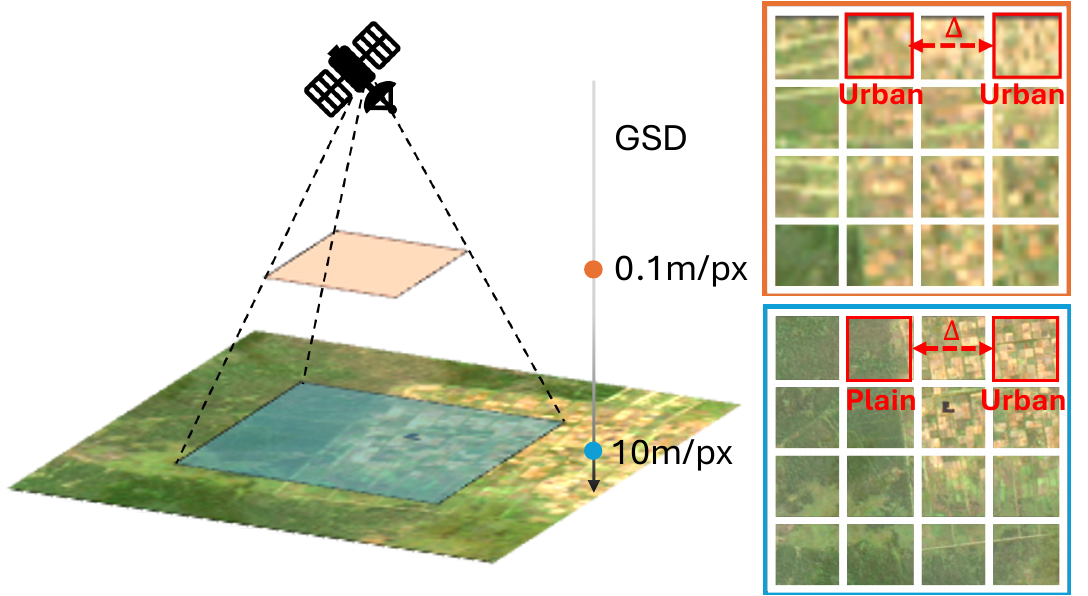}
\caption{
Motivation of GeoRoPE. 
The same token-grid size covers different ground extents under different GSDs. 
For a fixed token-grid offset $\Delta$, the corresponding ground distance and regional relation can also change, e.g., urban--urban at fine GSD versus plain--urban at coarse GSD. 
}
    \label{fig:motivation}
    \vspace{-10pt} 
\end{wrapfigure}

This issue is more fundamental in remote sensing than in natural-image recognition. 
In EO imagery, pixels and tokens are tied to physical ground measurements. 
As illustrated in Fig.~\ref{fig:motivation}, a fixed offset in image or token coordinates does not necessarily represent a fixed ground distance across sensors. 
When pretraining, fine-tuning or testing spans heterogeneous GSDs, the same grid-relative pattern may therefore carry conflicting physical meanings: as shown in Fig~\ref{fig:motivation}
a positional pattern that captures fine-grained local structures in high-resolution~(small GSD) imagery may instead cover much broader regions in lower-resolution imagery, leading to scale-misaligned attention responses.
Moreover, the meaning of a token offset is not determined by GSD alone. 
Remote-sensing scenes often contain regions with very different spatial granularity, ranging from expansive homogeneous areas to densely detailed urban landscapes. 
Therefore, the challenge is not only to align images across GSDs, but also to adapt token-level spatial relations to the ground distance they represent and to the spatial granularity of the observed regions.

Existing attempts to improve scale-robust representation learning in RSFMs mainly follow two directions. 
One direction incorporates physical-scale information into positional representations. 
For example, ScaleMAE~\cite{reed2023scale} introduces GSD-aware positional encoding to relate image coordinates to physical ground distances, encouraging more consistent representations across sensing resolutions. 
However, such physical alignment is typically applied as a uniform image-level prior: it anchors positions to ground distance, but does not adapt token-level positional interactions to regions with different spatial granularity. 
Another direction improves multi-scale spatial modeling through feature pyramids, hierarchical encoders, or parameter-efficient fine-tuning strategies~\cite{cong2022satmae, dong2024upetu,tang2023cross}. 
Related vision methods further use rotary position encoding to enhance the spatial sensitivity of attention~\cite{heo2024rotary,li2025cameras}. 
While these approaches enrich spatial representations, they either lack an explicit ground-distance anchor for token interactions or do not adapt positional sensitivity according to heterogeneous scene granularity. 
As a result, existing methods still lack a unified positional adaptation mechanism that can both ground token-grid interactions in physical distance and adjust their spatial sensitivity across different remote-sensing regions.

Therefore, we propose \textbf{GeoRoPE}, a RoPE-compatible and parameter-efficient spatial adaptation method for RSFMs. 
The key idea is to recalibrate token-level positional interactions from two complementary perspectives: how relative token offsets are measured in ground-aware units, and how strongly the model should respond to these offsets under different scene granularity. 
Specifically, \textit{Geo-Coordinate Calibration (GCC)} rescales raw token-grid offsets according to the ground distance represented by one token-grid step, producing geo-calibrated relative coordinates that are comparable across GSDs. 
Building on these calibrated coordinates, \textit{Geo-Frequency Calibration (GFC)} adjusts the native RoPE frequency with a bounded relation-specific factor, allowing positional sensitivity to vary across regions with different spatial granularity, such as compact built-up areas and smoother homogeneous landscapes. 
Together, GCC and GFC make RoPE-aware positional interactions sensitive to both cross-GSD ground-distance mismatch and intra-scene spatial heterogeneity. 
GeoRoPE is implemented as a lightweight attention-parallel adapter, which preserves the pretrained spatial prior of RSFMs while injecting a geo-aware positional correction with limited additional parameters.


The key contributions of this work are summarized as follows:
\begin{itemize}
    \item We propose \textbf{GeoRoPE}, a ground-aware spatial adaptation framework for RSFMs, to address the mismatch between token-grid positional interactions and real-world spatial relations under heterogeneous GSDs and scene granularity.
    
    \item We design RoPE-compatible calibration modules: \textit{Geo-Coordinate Calibration (GCC)}, which transforms index-based token offsets into ground-aware relative coordinates, and \textit{Geo-Frequency Calibration (GFC)}, which adaptively modulates positional sensitivity according to local scene granularity.
    
    \item Extensive evaluations across multiple RSFMs, sensors, and downstream tasks demonstrate that GeoRoPE consistently improves cross-resolution adaptation performance for both classification and segmentation.
\end{itemize}

\section{Related Work}
\label{sec:related work}

\noindent\textbf{Multi-Scale Representation in RSFMs.}
Remote sensing~(RS) data exhibits extreme physical heterogeneity, with varying ground sample distances~(GSDs) and sensor modalities making unified modeling highly nontrivial.
Early RSFMs~\cite{hong2023spectralgpt,sun2022ringmo,astruc2024omnisat} are typically pretrained at a \emph{fixed} input size or resolution, resizing all images to match the patch tokenizer. While effective, uniform scaling discards scale-specific cues. 
To mitigate this issue, recent works have explored multi-encoder architectures~\cite{guo2024skysense,han2024bridging,xiong2024one,astruc2025anysat} to capture features at modality- and scale-specific levels. While these strategies allow a single RSFM to process multi-resolution inputs, the resulting representations predominantly prioritize scale-invariant global robustness without addressing scale-dependent local structural fidelity.

\noindent\textbf{Adaptive Positional Encoding.}
Recent advancements have recontextualized positional encodings~\cite{vaswani2017attention, wu2021rethinking, su2024roformer} from static spatial identifiers to tunable scale-modeling mechanisms. 
Across both language and vision domains, most of existing methods adjust base wavelength of RoPE in LLMs~\cite{peng2023yarn,ding2024longrope}, or introduce adaptive phases in ViTs~\cite{heo2024rotary}, to achieve length or resolution extrapolation~\cite{wu2024efficient, zhu2023pose}. 
Beyond position extrapolation, other methods further inject explicit physical or geometric semantics into positional priors~\cite{li2025cameras}. PIPE~\cite{li2025pipe} incorporates latitude, longitude and day-of-year information for spatio-temporal forecasting, while Copernicus-FM~\cite{Wang_2025_ICCV} and Scale-MAE~\cite{reed2023scale} explicitly inject physical metadata and GSD into model with positional encodings. 
However, these methods fall short of addressing the unique demands of dense RS adaptation. The extreme intra-image structural heterogeneity inherent to RS data highlights the need for dynamically modulated, region-aware scale representations.

\noindent\textbf{Parameter-Efficient Fine-Tuning.}
A growing body of work adapts large pretrained backbones with a small number of task-specific parameters.
Representative approaches, including bias or normalization term tuning~\cite{zaken2022bitfit, zhao2023tuning}, adapters~\cite{houlsby2019parameter} and Low-Rank Adaptation~(LoRA)~\cite{hu2022lora}, have demonstrated an effective trade-off between adaptation flexibility and computational efficiency. 
In remote-sensing field, recent PEFT studies have mainly focused on mitigating modality discrepancies. 
For example, Scaled Low-Rank Adaptation~\cite{scheibenreif2024parameter} rescales LoRA updates to improve transfer to unseen modalities, while SpectralX~\cite{zhang2025spectralx} and DEFLECT~\cite{thoreau2025parameter} disentangle spectral and spatial adaptation branches to bridge RGB, multispectral, and hyperspectral imagery. 
Despite their effectiveness for modality transfer, these PEFT methods generally treat spatial adaptation as a static process. As a result, how to realize dynamically modulated, region-aware scale adaptation within a lightweight PEFT framework remains underexplored. 


\section{Methodology}
\label{sec:method}

\subsection{Problem Formulation}
\label{sec:preliminary:problem}

We consider a remote-sensing observation $\mathbf{X} \in \mathbb{R}^{T \times H \times W \times C}$ acquired at a known GSD $G$, where $G$ denotes the physical length represented by one image pixel. Here, $T$ is the temporal length, $H$ and $W$ are spatial dimensions, and $C$ is the number of spectral channels. 
In typical RSFMs, temporal information, when present, is aggregated before the main spatial attention blocks. Accordingly, the core attention blocks encode $\mathbf X$ into a spatial token map $\mathbf Z^{(l)} \in \mathbb R^{N_x^{(l)}\times N_y^{(l)}\times D_l}$ across different stages $l$, where $N_x^{(l)}$ and $N_y^{(l)}$ are the spatial token-grid dimensions, and $D_l$ is the feature dimension. 
Each token is indexed by a 2D grid coordinate $\mathbf u=(u_x,u_y)\in\mathcal N^{(l)}$, where $\mathcal N=\{0,\ldots,N_x^{(l)}-1\}\times\{0,\ldots,N_y^{(l)}-1\}$. 
For two tokens $\mathbf u,\mathbf v\in\mathcal N^{(l)}$, their signed relative grid offset is:
\begin{equation}
    \boldsymbol{\Delta}_{\mathbf u,\mathbf v}
    =
    \mathbf u-\mathbf v =  (\Delta^x_{\mathbf u,\mathbf v},  \Delta^y_{\mathbf u,\mathbf v})
    =
    (u_x-v_x,\;u_y-v_y).
\end{equation}
where the stage index $l$ is omitted for simplicity.

Spatial Transformers in RSFMs typically represent positional relations in token-grid units. 
However, such grid-level relations cannot reflect the physical distance on the ground. 
Given a token map corresponds to a $\tau \times \tau$ image patch acquired at GSD $G$, its physical side length is:

\begin{equation}
    \ell = \tau G .
\end{equation}

Under this definition, the physical distance between two token is:
\begin{equation}
\label{eq:physical-displacement}
    \mathbf d_{\mathbf u,\mathbf v}
    =
    \ell
    \boldsymbol{\Delta}_{\mathbf u,\mathbf v}.
\end{equation}

\textbf{Motivations.} 
This geometry reveals a scale ambiguity in token-grid positional relations. 
A fixed offset $\boldsymbol{\Delta}_{\mathbf u,\mathbf v}$ does not necessarily imply a fixed physical displacement, since the physical meaning of a token-grid unit changes with the physical token scale. 
When training or fine-tuning spans heterogeneous GSDs, the same grid-relative pattern may correspond to different ground extents, inducing conflicting physical interpretations for the same grid-based positional prior. 
When transferred to a new physical token scale, this grid-based prior may be reused at a mismatched ground extent, leading to scale-misaligned attention responses. 
This motivates revisiting the positional formulation to calibrate grid-relative offsets into physically meaningful relative coordinates for RSFMs.

\subsection{Rotary Position Embedding on Spatial Token Grids}
\label{sec:preliminary:rope}

We now revisit how such offsets influence standard spatial attention with Rotary Position Embedding (RoPE)~\cite{su2024roformer}
RoPE encodes relative positions by applying coordinate-dependent rotations to query and key vectors in Transformers. Let 
$\mathbf Q_{\mathbf u}^{(h)}, \mathbf K_{\mathbf v}^{(h)} \in \mathbb R^{d_r}$ denote the query and key vectors at head $h$, where $d_r$ is the number of per-head channels to which RoPE is applied. 
For 2D RoPE, $d_r$ is divisible by four and set $M=d_r/4$. 
The base angular frequencies are defined as:
\begin{equation}
\label{eq:rope-base}
    \theta_m
    =
    \mathrm{base}^{-4m/d_r},
    \qquad
    m=0,\ldots,M-1,
\end{equation}
where $\mathrm{base}$ is commonly set to $10{,}000$.

For a scalar phase $\phi$, the elementary rotation block is defined as:
\begin{equation}
\label{eq:rope-2d-block}
    \mathbf r(\phi)
    =
    \begin{bmatrix}
    \cos \phi & -\sin \phi \\
    \sin \phi & \cos \phi
    \end{bmatrix}.
\end{equation}
Given the relative grid offset  $\boldsymbol{\Delta}_{\mathbf u,\mathbf v}$,
the full equation of 2D relative RoPE rotation matrix~\cite{heo2024rotary} is:
\begin{equation}
\label{eq:rope-rotation}
\mathbf R_{\boldsymbol{\Delta}_{\mathbf u,\mathbf v}}
=
\operatorname{blockdiag}
\left(
\mathbf r(\theta_0\Delta^x_{\mathbf u,\mathbf v}),
\mathbf r(\theta_0\Delta^y_{\mathbf u,\mathbf v}),
\ldots,
\mathbf r(\theta_{M-1}\Delta^x_{\mathbf u,\mathbf v}),
\mathbf r(\theta_{M-1}\Delta^y_{\mathbf u,\mathbf v})
\right).
\end{equation}

Accordingly, the RoPE-modulated attention logit can be written as:
\begin{equation}
\label{eq:relative-property}
\mathbf a_{\mathbf u,\mathbf v}^{(l,h)}
=
(\widetilde{\mathbf Q}_{\mathbf u}^{(h)})^\top
\widetilde{\mathbf K}_{\mathbf v}^{(h)}
=
(\mathbf Q_{\mathbf u}^{(h)})^\top
\mathbf R_{\boldsymbol{\Delta}_{\mathbf u,\mathbf v}}
\mathbf K_{\mathbf v}^{(h)},
\end{equation}

\textbf{Scale implication of RoPE.}
Standard RoPE assigns a fixed angular-frequency schedule to discrete token-grid offsets. 
Although this design provides a stable relative positional prior in grid coordinates, its physical meaning changes when the ground extent represented by one token step changes. 
In remote sensing, the same rotary channel may correspond to fine-grained structures under a high-resolution GSD but to broader ground regions under a coarser GSD. 
When learning spans heterogeneous GSDs, this can associate the same grid-based positional prior with inconsistent physical meanings. 
When the model is adapted or transferred to imagery with a different physical token scale, the learned prior may further produce scale-misaligned attention responses. 
This motivates recalibrating the coordinate unit of RoPE: the offset driving that phase should be aware of the physical token scale.

\begin{figure}[t]
    \centering 
    \includegraphics[width=\linewidth]{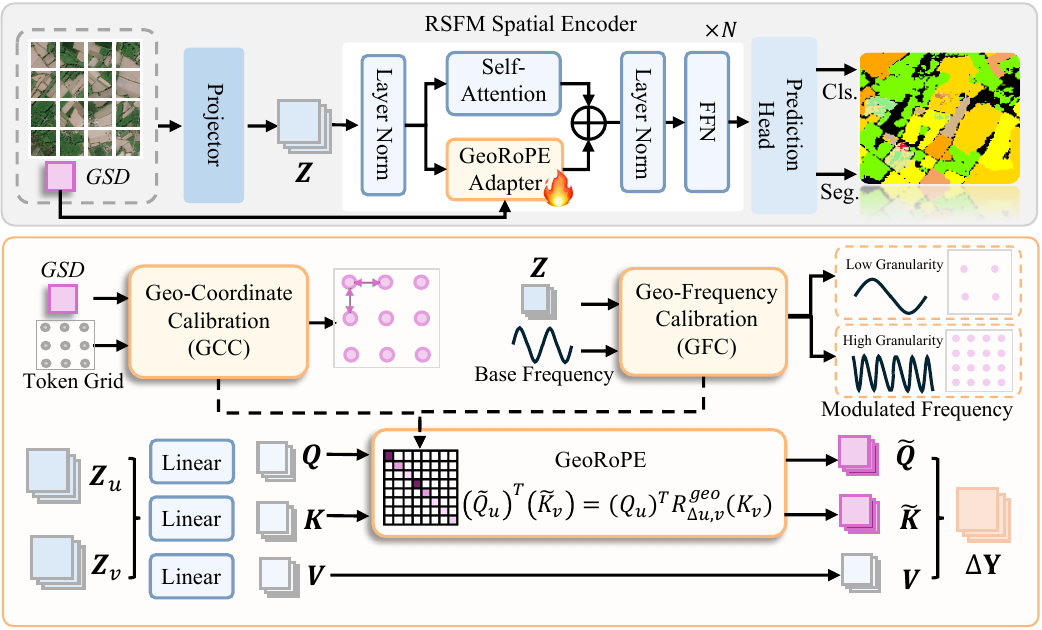}
        \caption{
Overview of the GeoRoPE framework. 
GeoRoPE is injected into RSFM Transformer blocks as a lightweight attention-parallel adapter, while the pretrained backbone path is preserved for downstream classification and segmentation. 
Within the adapter, Geo-Coordinate Calibration (GCC) converts raw token-grid offsets into ground-aware relative coordinates, and Geo-Frequency Calibration (GFC) adjusts positional sensitivity according to scene-dependent spatial granularity. 
The resulting GeoRoPE rotary relation $\mathbf R^{\mathrm{geo}}_{\mathbf{\Delta}_{\mathbf u,\mathbf v}}$ is applied to the adapter queries and keys, producing an additive geo-aware correction $\Delta\mathbf Y$ to the backbone attention output.
}
\label{fig:model_figure}
\end{figure}

\subsection{GeoRoPE: A Unified RS Scale Modeling Framework}
\label{sec:georope}

\subsubsection{Overview}
\label{sec:georope:overview}

We propose GeoRoPE, a scale-aware rotary adaptation mechanism that recalibrates the coordinate unit driving the RoPE phase. 
Rather than replacing the rotary structure, GeoRoPE preserves the pairwise rotary block form of RoPE while making the relative phase aware of the physical token scale and local scene context. 

For a token pair $\mathbf u,\mathbf v$, standard RoPE forms its phase from the product between a fixed frequency $\theta_m$ and the raw grid offset $\boldsymbol{\Delta}_{\mathbf u,\mathbf v}$. 
GeoRoPE calibrates the RoPE phase from by introducing two complementary modules:
First, Geo-Coordinate Calibration (GCC) rescales the relative grid offset by a physical scale factor $\rho_G$, converting raw token-grid steps into geo-calibrated coordinate units. 
Second, Geo-Frequency Calibration (GFC) adjusts the native RoPE frequency $\theta_m$ by a bounded relation-specific factor 
$\alpha_{\mathbf u,\mathbf v,m}$, controlling how rapidly the rotary phase varies over the geo-calibrated coordinate. 
This allows RoPE to adapt its positional sensitivity to scene-dependent spatial granularity while remaining bounded and RoPE-compatible. 

Overall, the GeoRoPE phase at frequency $m$ can be written compactly as:
\begin{equation}
\label{eq:georope-decomposition}
    \boldsymbol{\phi}^{\mathrm{geo}}_{\mathbf u,\mathbf v,m}
    =
    \underbrace{
    \theta_m
    \alpha_{\mathbf u,\mathbf v,m}
    }_{
    \substack{
    \text{GFC: granularity-aware}\\
    \text{phase variation rate}
    }}
    \;\cdot\;
    \underbrace{
    \rho_G
    \boldsymbol{\Delta}_{\mathbf u,\mathbf v}
    }_{
    \substack{
    \text{GCC: physical-scale}\\
    \text{coordinate unit}
    }}
    .
\end{equation}



The resulting GeoRoPE relative rotation matrix is:
\begin{equation}
\label{eq:georope-matrix}
\mathbf R^{\mathrm{geo}}_{\Delta_{\mathbf u,\mathbf v}}
=
\operatorname{blockdiag}
\left(
\mathbf r(\phi^{\mathrm{geo},x}_{\Delta_{\mathbf u,\mathbf v},0}),
\mathbf r(\phi^{\mathrm{geo},y}_{\Delta_{\mathbf u,\mathbf v},0}),
\ldots,
\mathbf r(\phi^{\mathrm{geo},x}_{\Delta_{\mathbf u,\mathbf v},M-1}),
\mathbf r(\phi^{\mathrm{geo},y}_{\Delta_{\mathbf u,\mathbf v},M-1})
\right),
\end{equation}
where $\mathbf r(\cdot)$ is the standard $2\times2$ rotary block and $M=d_r/4$ is the number of frequency bands per spatial axis. 
When $\rho_G=1$ and $\alpha_{\mathbf u,\mathbf v,m}=1$ for all token pairs and frequency bands, GeoRoPE reduces to the original grid-based RoPE.


\subsubsection{Geo-Coordinate Calibration}\label{sec:method:global}

Geo-Coordinate Calibration (GCC) calibrates the coordinate unit by expressing token-grid offsets relative to a reference ground distance. 
Let $\ell_{\mathrm{ref}}$ denote the reference ground distance, chosen by default as the finest resolution such distance among the training inputs. 
Given the current token-step ground distance $\ell$, GCC defines the coordinate-scale factor as:
\begin{equation}
\label{eq:global-scale}
    \rho_G
    =
    \left(
    \frac{\ell}{\ell_{\mathrm{ref}}}
    \right)^{1-\gamma},
    \qquad
    \gamma\in[0,1].
\end{equation}
The geo-calibrated relative coordinate is then defined as:
\begin{equation}
\label{eq:calibrated-offset}
    \widetilde{\boldsymbol{\Delta}}_{\mathbf u,\mathbf v}
    =
    \rho_G
    \boldsymbol{\Delta}_{\mathbf u,\mathbf v}.
\end{equation}


The exponent $\gamma$ controls how strongly the RoPE coordinate unit is tied to the ground distance represented by one token-grid step. 
When $\gamma=1$, $\rho_G=1$, and GCC reduces to the original grid-based coordinate system. 
When $\gamma=0$, offsets are measured in units of the reference token-step ground distance $\widetilde{\boldsymbol{\Delta}}_{\mathbf u,\mathbf v}
=(\ell/\ell_{\mathrm{ref}})\boldsymbol{\Delta}_{\mathbf u,\mathbf v}$. 
In this case, token pairs with the same ground displacement produce the same calibrated coordinate, even if their raw grid offsets differ under different GSDs. 

This tempering helps preserve the stability of pretrained RoPE. 
Under strict calibration, a large ratio 
$\ell^{(l)}/\ell_{\mathrm{ref}}^{(l)}$ 
can substantially enlarge the rotary phase, especially in high-frequency channels. 
This may cause excessive phase wrapping and disturb the grid-based positional prior inherited from a pretrained RSFM. 
Thus, GCC interpolates between two coordinate regimes: the original grid-relative coordinate system and a ground-calibrated coordinate system.




\subsubsection{Geo-Frequency Calibration}
\label{sec:method:modulation}

Geo-Coordinate Calibration aligns RoPE with the physical resolution of the input, but a single global scale factor cannot capture the spatial heterogeneity within remote-sensing scenes. 
Dense urban regions, homogeneous croplands, and fragmented natural landscapes may require different positional sensitivities even under the same GSD.

To account for this heterogeneity, Geo-Frequency Calibration (GFC) adjusts the phase variation rate of RoPE over the geo-calibrated coordinate, which predicts a spatially-varying modulation factor $\alpha_{\mathbf{u},m}$ for token location $\mathbf{u}$ and frequency index $m$. 
Formally, given the input spatial token map $\mathbf{Z} \in \mathbb{R}^{N_x \times N_y \times D}$, the modulation is formulated as:

\begin{equation}
\label{eq:token-modulation}
    \alpha_{\mathbf{u},m}
    =
    \exp
    \Big(
    \tanh \big( \mathcal{F}_m(\mathbf{Z}_{\mathbf{u})} \big)
    \Big).
\end{equation}
where $\mathcal{F}$ is a semantic extraction network. The $\tanh$ nonlinearity bounds the factor within $(-1, 1)$, while the exponential map keeps the multiplier positive:
$\alpha_{\mathbf u,m}\in(e^{-1},e)$.

Specifically, the extraction network $\mathcal{F}$ maps the token embedding $\mathbf{Z}_{\mathbf{u}}$ into a frequency-modulating latent space $\mathbf{H} \in \mathbb{R}^{N_x \times N_y \times M}$, based on a fine-grained local texture descriptor $\mathbf{F}_l$ and a holistic global landscape prior $\mathbf{F}_g$:

\begin{equation}
\label{eq:asgm-fusion}
    \mathbf H
    =
    \mathrm{Conv}_{1\times1}
    \left(
    [\mathbf F_l,\mathrm{Broadcast}(\mathbf F_g)]
    \right)
    \in
    \mathbb R^{N_x\times N_y\times M},
\end{equation}
where
\begin{equation}
\label{eq:asgm-feature}
    \mathbf F_l
    =
    \mathrm{GELU}
    \left(
    \mathrm{Conv}_{\mathrm{DW}}(\mathbf Z)
    \right),
    \qquad
    \mathbf F_g
    =
    \mathrm{MLP}
    \left(
    \mathrm{GAP}(\mathbf Z)
    \right).
\end{equation}

\textbf{Pair-wise frequency calibration.}
Since GeoRoPE is applied to token pairs, We aggregate them symmetrically in the log domain: $\alpha_{\mathbf u,\mathbf v,m}=\sqrt{\alpha_{\mathbf u,m}\alpha_{\mathbf v,m}}$. It rescales the relative GeoRoPE phase, thereby adapting the positional sensitivity across homogeneous regions.

\subsection{Adapter for GeoRoPE Injection}
\label{sec:method:peft}

To integrate GeoRoPE without disrupting the pretrained spatial prior, we inject it through a lightweight attention-parallel adapter, with the native attention and positional encoding path of the pretrained RSFM frozen.

Given input tokens $\mathbf Z$, the frozen backbone yields the original attention output $\mathbf Y_{\mathrm{frozen}}$, while the trainable branch computes a feature-level correction in a reduced latent space:
\begin{equation}
\label{eq:geo-adapter-correction}
    \Delta \mathbf Y
    = 
    \mathrm{Attn}_{\mathrm{GeoRoPE}} 
    \left( 
    \mathbf Z; 
    \mathbf R^{\mathrm{geo}} 
    \right),
\end{equation}
where $\mathbf R^{\mathrm{geo}}$ is the GeoRoPE-calibrated rotary relation. The final block output is obtained via residual addition:
\begin{equation}
\label{eq:adapter-residual}
    \mathbf Y 
    = 
    \mathbf Y_{\mathrm{frozen}} 
    + 
    \Delta \mathbf Y^.
\end{equation}
This parameter-efficient feature-level injection provides an additive geospatial correction without overriding the backbone's learned representations.



\section{Experiment}

\subsection{Experimental Settings}

\noindent\textbf{Datasets and metrics.} 
Our evaluation is conducted on {GeoBench}~\cite{lacoste2023geo}, including six \textbf{classification} datasets: m-bigearthnet~(BE), m-brick-kiln~(BK), m-eurosat~(ES), m-forestnet~(FN), m-pv4ger~(PV), m-so2sat~(SS), and six \textbf{semantic segmentation} datasets: m-cashew-plantation~(CW), m-NeonTree~(NT), m-nz-cattle~(CT), m-chesapeake-landcover~(CP), m-pv4ger-seg~(PS) and m-SA-crop-type~(CV), spanning ground sample distances from $0.1$\,m (RGB aerial) to $15$\,m (Landsat-8). 
For classification, we report the mean Accuracy score. For segmentation, we report Intersection over Union (IoU) for 2-classes datasets and mean IoU (mIoU) otherwise.
More details are in Appendix~\ref{appendix:dataset_details}.

\noindent\textbf{Backbones and baselines.}
We evaluate GeoRoPE along two complementary dimensions. 
First, to examine its compatibility with different pretrained priors, we instantiate GeoRoPE on representative RSFMs with diverse native positional encodings: 
\textbf{DOFA}~\cite{xiong2024neural} with absolute positional encoding~(APE), 
\textbf{Scale-MAE}~\cite{reed2023scale} with GSD-aware positional encoding, and 
\textbf{SatLas}~\cite{bastani2023satlaspretrain} with relative positional encoding~(RPE) for heterogeneous EO inputs. 
Across all backbones, GeoRoPE is attached through the same attention-parallel rotary adapter and does not require backbone-specific redesign. 
Second, to assess GeoRoPE as a parameter-efficient adaptation method, we compare it with representative PEFT baselines, including BitFit~\cite{zaken2022bitfit}, NormTuning~\cite{zhao2023tuning}, LoRA~\cite{hu2022lora}, and Scaled Low-Rank (SLR)~\cite{scheibenreif2024parameter}. 
For all LoRA-based methods, the rank is set to $8$ following the common setting in~\cite{hu2022lora}. 
More details are provided in Appendix~\ref{appendix:backbone_details}.



\noindent\textbf{Implementation details.} 
Following the base model settings, 
for classification, we follow a linear probing protocol, in which a \texttt{[CLS]} token is prepended to the average-pooled token embeddings for prediction; for segmentation, we use the UperNet~\cite{xiao2018unified} as decoder on top of the frozen pretrained backbone. 
The total number of training epochs is fixed to 100 for each dataset and GeoRoPE is optimized in the attention layers, following the standard PEFT setup.
More details are in Appendix~\ref{appendix:training_details}


\begin{table*}[t]
    \centering
    \caption{
    \textbf{Comparison of performances on classification and segmentation tasks.}
    Avg.C, Avg.S, and Avg.P denote the macro-average performance over classification datasets, segmentation datasets, and all datasets, respectively.
    Bold indicates the best performance within each backbone group.
    }
    \label{tab:main_results}

    \resizebox{\textwidth}{!}{%
    \begin{tabular}{l cccccc cccccc ccc}
        \toprule
        \multirow{2}{*}{\textbf{Method}} 
        & \multicolumn{6}{c}{\textbf{Classification}} 
        & \multicolumn{6}{c}{\textbf{Segmentation}} 
        & \multicolumn{3}{c}{\textbf{Average}} \\
        \cmidrule(lr){2-7} \cmidrule(lr){8-13} \cmidrule(lr){14-16}
        & BE & BK & ES & FN & PV & SS 
          & CW & NT & CT & CP & PS & SC 
          & Avg.C & Avg.S & Avg.P \\
        \midrule

        \multicolumn{16}{c}{\textbf{ScaleMAE}} \\
        \midrule
        LP/UP 
        & \textbf{72.50} & 84.78 & 72.76 & 33.98 & 87.60 & 32.96
        & 58.34 & 56.96 & 72.60 & 53.47 & 90.63 & \textbf{29.16}
        & 64.10 & 60.19 & 62.15 \\

        BitFit 
        & 66.39 & 96.39 & 91.75 & 35.78 & 91.19 & 47.16
        & 58.74 & 55.27 & 72.60 & 55.49 & 90.88 & 29.43
        & 71.44 & 60.40 & 65.92 \\

        NormTuning 
        & 68.23 & 91.69 & 79.11 & 35.03 & 90.49 & 40.97
        & 58.48 & 55.33 & 72.83 & 53.08 & 90.72 & 28.54
        & 67.59 & 59.83 & 63.71 \\

        LoRA 
        & 68.37 & 97.49 & 94.73 & 46.71 & 95.29 & 47.57
        & 59.96 & 55.09 & 73.24 & 61.73 & \textbf{92.55} & 25.79
        & 75.03 & 61.39 & 68.21 \\

        SLR 
        & 68.12 & 97.20 & 96.05 & \textbf{47.75} & 94.90 & \textbf{53.04}
        & 59.32 & 54.59 & 73.26 & \textbf{65.12} & 92.48 & 26.31
        & 76.06 & 61.85 & 68.95 \\

        \rowcolor{gray!10}
        \textbf{GeoRoPE (ours)} 
        & 69.18 & \textbf{98.00} & \textbf{96.67} & 46.59 & \textbf{95.70} & \textbf{53.04}
        & \textbf{61.23} & \textbf{57.21} & \textbf{73.43} & 61.86 & 92.43 & 26.49
        & \textbf{76.53} & \textbf{62.11} & \textbf{69.32} \\

        \midrule
        \multicolumn{16}{c}{\textbf{SatLas}} \\
        \midrule
        LP/UP
        & 69.94 & 83.35 & 73.29 & 46.71 & 93.84 & 43.00
        & 43.80 & 57.62 & 83.26 & 63.46 & 92.73 & 33.76
        & 68.35 & 62.44 & 65.98 \\

        BitFit 
        & 70.43 & 97.39 & 98.63 & 53.14 & 97.39 & 52.43
        & 66.86 & 58.02 & 83.74 & 65.75 & 94.11 & 34.68
        & 78.24 & 67.19 & 72.71 \\

        NormTuning 
        & 69.58 & 97.20 & 98.43 & 50.00 & 97.30 & 53.35
        & 65.71 & 57.59 & 83.50 & 65.43 & 93.57 & 33.55
        & 77.64 & 66.56 & 72.10 \\

        LoRA 
        & 70.42 & 97.80 & \textbf{98.79} & 53.89 & 97.69 & \textbf{60.45}
        & 68.99 & 57.93 & 84.23 & 68.68 & 94.82 & 35.52
        & 79.84 & 68.36 & 74.10 \\

        SLR 
        & 71.40 & 97.80 & 98.70 & 53.59 & \textbf{98.00} & 59.94
        & 69.38 & 57.95 & \textbf{84.28} & 67.70 & 94.85 & 35.51
        & \textbf{79.91} & 68.28 & 74.09 \\

        \rowcolor{gray!10}
        \textbf{GeoRoPE (ours)} 
        & \textbf{71.48} & \textbf{97.98} & 98.40 & \textbf{54.04} & 97.90 & 56.40
        & \textbf{70.40} & \textbf{58.48} & \textbf{84.28} & \textbf{68.75} & \textbf{95.00} & \textbf{36.56}
        & 79.37 & \textbf{68.91} & \textbf{74.14} \\

        \midrule
        \multicolumn{16}{c}{\textbf{DoFA}} \\
        \midrule
        LP/UP
        & 87.90 & 96.80 & 87.00 & 41.47 & 96.99 & 55.10
        & 61.20 & 57.40 & 73.80 & 56.92 & 92.60 & 31.00
        & 77.54 & 62.15 & 69.85 \\

        BitFit 
        & 89.41 & 98.30 & 95.82 & 50.60 & 97.99 & 57.20
        & 67.06 & 55.64 & 74.99 & 59.23 & 94.03 & 33.48
        & 81.55 & 64.07 & 72.81 \\

        NormTuning 
        & 89.45 & 98.49 & 94.04 & 52.44 & 98.10 & 58.62
        & 66.38 & 55.49 & 74.78 & 58.39 & 93.81 & 32.88
        & 81.86 & 63.62 & 72.74 \\

        LoRA 
        & 88.13 & \textbf{98.50} & 96.92 & \textbf{54.24} & 97.99 & 60.53
        & \textbf{68.61} & 55.97 & 75.26 & 60.56 & 94.14 & 33.57
        & 82.72 & 64.69 & 73.70 \\

        SLR 
        & 88.86 & 98.30 & 97.20 & 53.74 & 98.00 & 61.56
        & 68.19 & 56.16 & 75.29 & 60.34 & 94.10 & 34.10
        & 82.94 & 64.70 & 73.82 \\

        \rowcolor{gray!10}
        \textbf{GeoRoPE (ours)} 
        & \textbf{94.16} & 98.48 & \textbf{97.44} & 54.21 & \textbf{98.21} & \textbf{62.78}
        & 68.48 & \textbf{59.19} & \textbf{75.45} & \textbf{61.54} & \textbf{94.39} & \textbf{35.19}
        & \textbf{84.21} & \textbf{65.71} & \textbf{74.96} \\

        \bottomrule
    \end{tabular}%
    }
\vspace{-1.2em}
\end{table*}

\subsection{Overall Performance}

\noindent\textbf{Quantitative results.}
Table~\ref{tab:main_results} reports the results on three RSFMs with different native positional encodings. 
Overall, GeoRoPE achieves the best Avg.P under all three backbones, indicating that the proposed geo-aware rotary adapter can complement diverse pretrained RSFMs without backbone-specific redesign.
For \textbf{ScaleMAE}, GeoRoPE improves Avg.C, Avg.S, and Avg.P to 76.53, 62.11, and 69.32, respectively, outperforming the strongest PEFT baseline SLR on all averaged metrics. 
This suggests that explicitly adapting positional interactions is more effective than generic parameter-efficient updates under the same backbone. 
Notably, since ScaleMAE already incorporates GSD-aware positional information, the additional gains indicate that GeoRoPE can complement existing scale-aware positional priors rather than merely replacing them.
For \textbf{Satlas}, GeoRoPE achieves the best Avg.S and Avg.P, reaching 68.91 and 74.14. 
Although its Avg.C is slightly lower than LoRA and SLR, it improves the overall average by producing stronger segmentation performance, which is more directly tied to token-level spatial alignment. 
For \textbf{DoFA}, GeoRoPE obtains the best Avg.C, Avg.S, and Avg.P, reaching 84.21, 65.71, and 74.96, respectively, showing the clearest overall gains among the evaluated backbones.

Across task types, GeoRoPE is particularly consistent on \textbf{segmentation}, where it achieves the highest Avg.S for all three backbones. 
This is aligned with the motivation of GeoRoPE: dense prediction is more sensitive to whether token-grid relations correspond to meaningful ground distances and appropriate positional sensitivities. 
For \textbf{classification}, GeoRoPE remains best or highly competitive, but the gains are less uniform, as global recognition can rely more heavily on semantic aggregation than fine-grained positional calibration. 
These results suggest that GeoRoPE is most effective when spatial correspondence is critical, while still maintaining strong general-purpose adaptation.


\begin{figure}[t]
    \centering 
    \includegraphics[width=\linewidth]{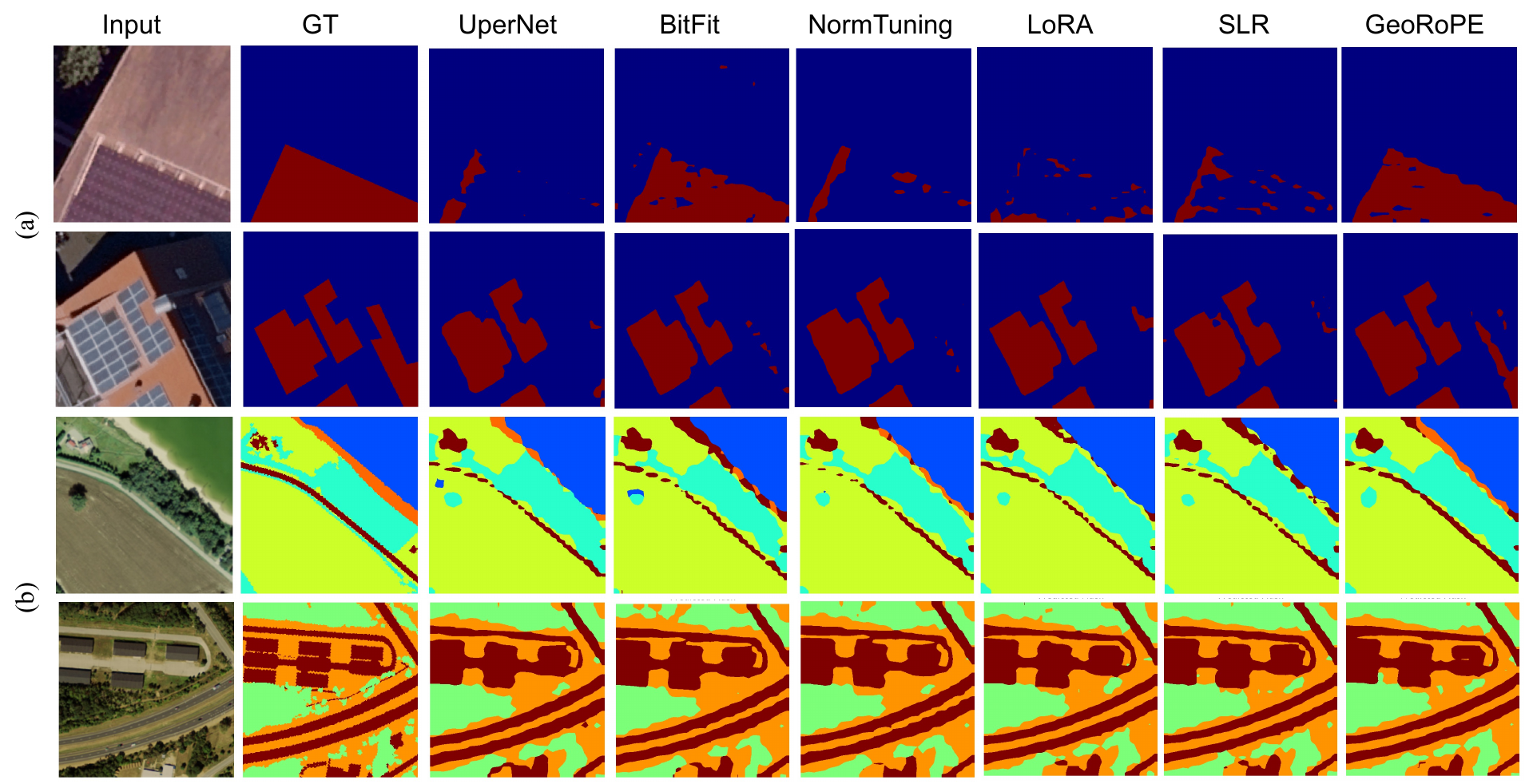}
        \caption{Segmentation with different PEFT methods on imagery from (a) \textit{pv4ger} and (b) \textit{chesapeake}.}   
\vspace{-1.5em}
\label{fig:segmentation}
\end{figure}


\noindent\textbf{Qualitative results.}
Fig.~\ref{fig:segmentation} compares segmentation outputs on PV4GER and Chesapeake. 
On PV4GER, GeoRoPE produces more complete rooftop solar-panel masks with fewer fragmented or missing foreground regions. 
On Chesapeake, it yields more coherent land-cover regions and cleaner class boundaries than generic PEFT strategies. 
These visual results indicate that calibrating both ground-aware coordinate units and granularity-aware positional sensitivity helps pretrained RSFMs maintain spatially coherent predictions across heterogeneous remote-sensing scenes. More visualizations are provided in Appendix~\ref{appendix:visualizaitons}


\newcolumntype{C}[1]{>{\centering\arraybackslash}p{#1}}
\newcolumntype{L}[1]{>{\raggedright\arraybackslash}p{#1}}

\begin{wraptable}{r}{0.5\textwidth}
    \vspace{-1.4em}
    \centering
    \caption{
    \textbf{Ablation study of GeoRoPE.}
    We progressively add GGC, GFC, and the adapter injection~(Inj.) branch.
    }
    \label{tab:ablation_georope}
    \small
    \setlength{\tabcolsep}{2.2pt}
    \renewcommand{\arraystretch}{1.05}

    \begin{tabular}{L{0.18\linewidth} C{0.08\linewidth} C{0.08\linewidth} C{0.08\linewidth} C{0.105\linewidth} C{0.105\linewidth} C{0.105\linewidth}}
    \toprule
    \textbf{Method} & \textbf{GGC} & \textbf{GFC} & \textbf{Inj.} & \textbf{Cls.} & \textbf{Seg.} & \textbf{Avg.} \\
    \midrule
    LoRA 
    & $\times$ & $\times$ & $\times$ 
    & 82.72 & 64.69 & 73.70 \\

    RoPE 
    & $\times$ & $\times$ & $\checkmark$ 
    & 82.58 & 64.72 & 73.65 \\

    + GGC 
    & $\checkmark$ & $\times$ & $\checkmark$ 
    & 82.98 & 64.87 & 73.93 \\

    + GFC
    & $\times$ & $\checkmark$ & $\checkmark$ 
    & 82.66 & 64.89 & 73.78 \\

    \midrule
    \rowcolor{gray!10}
    \textbf{GeoRoPE} 
    & $\checkmark$ & $\checkmark$ & $\checkmark$ 
    & \textbf{84.21} & \textbf{65.71} & \textbf{74.96} \\
    \bottomrule
    \end{tabular}
    \vspace{-1.2em}
\end{wraptable}


\subsection{Ablation Study and Sensitivity Analysis}

We compare the standard LoRA baseline, RoPE injection without geo-coordinate calibration~(GCC), geo-frequency calibration~(GFC), and the full GeoRoPE.
All reported results are averaged over all classification and segmentation benchmarks.

\noindent\textbf{Effectiveness of GeoRoPE injection.} 
A plain RoPE injection branch does not improve over LoRA, obtaining 73.65 Avg. compared with 73.70 for LoRA. 
Although it slightly improves segmentation, its classification score changes from 82.72 to 82.58, showing no consistent gain across tasks. 
This indicates that standard RoPE is insufficient for robust scale-aware adaptation. 
Without explicit geo-coordinate and geo-frequency calibration, the rotary positional signal remains tied to token-grid coordinates rather than ground-aware spatial relations.
In contrast, full GeoRoPE reaches 74.96 Avg., surpassing both LoRA and plain RoPE injection. 
This confirms that the gains come from geo-aware positional calibration rather than merely from adding an extra adapter branch.

\begin{figure*}[ht]
    \centering
    \renewcommand{\arraystretch}{1.0}
    \vspace{-0.8em}
    \begin{subfigure}[t]{0.32\textwidth}
        \centering
        \caption{GCC calibration.}
        \label{tab:ablation_geocalibration}
        \vspace{0.3em}




\scriptsize
\setlength{\tabcolsep}{2.2pt}
\renewcommand{\arraystretch}{1.0}
\resizebox{\linewidth}{!}{%
\begin{tabular}{lccc}
\toprule
\textbf{Calibration} & \textbf{Cls.} & \textbf{Seg.} & \textbf{Avg.} \\
\midrule
No calibration        & 82.66 & 64.89 & 73.78 \\
Physical scaling      & 82.62 & 64.77 & 73.70 \\
Log-space calibration & 83.02 & 65.47 & 74.24 \\
\midrule
\rowcolor{gray!10}
\textbf{GeoRoPE} & \textbf{84.21} & \textbf{65.71} & \textbf{74.96} \\
\bottomrule
\end{tabular}
}
    \end{subfigure}
    \hfill
    \begin{subfigure}[t]{0.32\textwidth}
        \centering
        \caption{GFC modulation.}
        \label{tab:ablation_frequency_modulation}
        \vspace{0.3em}




\scriptsize
\setlength{\tabcolsep}{2.2pt}
\renewcommand{\arraystretch}{1.0}
\resizebox{\linewidth}{!}{%
\begin{tabular}{lccc}
\toprule
\textbf{Calibration} & \textbf{Cls.} & \textbf{Seg.} & \textbf{Avg.} \\
\midrule
No calibration     & 82.98 & 64.87 & 73.93 \\
Static scaling & 82.69    & 65.21    & 73.95    \\
\midrule
\rowcolor{gray!10}
\textbf{GeoRoPE} & \textbf{84.21} & \textbf{65.71} & \textbf{74.96} \\
\bottomrule
\end{tabular}
}
    \end{subfigure}
    \hfill
    \begin{subfigure}[t]{0.32\textwidth}
        \centering
        \caption{Sensitivity to $\gamma$.}
        \label{fig:gamma}
        \vspace{0.3em}
        \includegraphics[width=\linewidth]{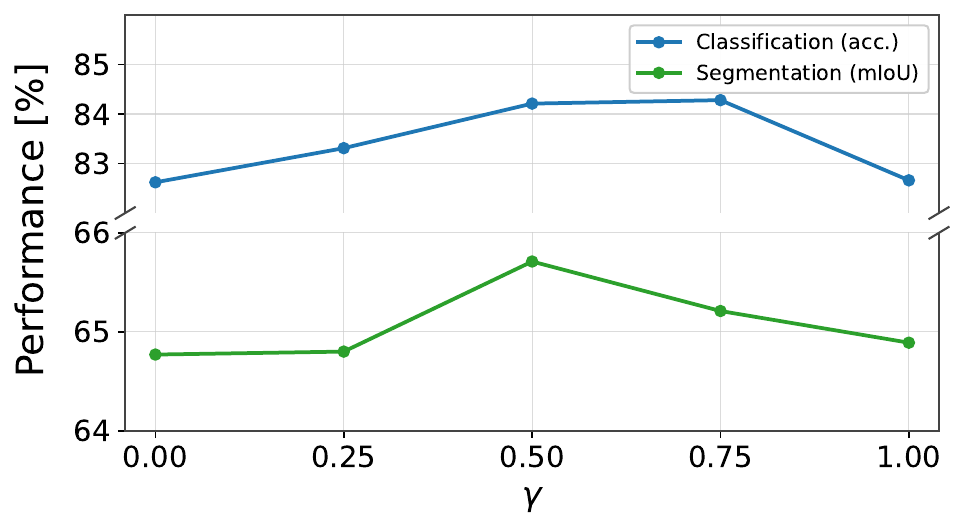}
    \end{subfigure}
    \vspace{-0.4em}
    \caption{
    \textbf{Analysis of GeoRoPE components and modulation strength.}
    We compare GCC calibration, GFC frequency modulation, and the sensitivity of GeoRoPE to $\gamma$.}
    \label{fig:ablation_components}
    \vspace{-0.8em}
\end{figure*}

\noindent\textbf{Effectiveness of GCC Module.} 
Adding GCC to the plain RoPE injection branch improves the average performance from 73.65 to 73.93, indicating that calibrating the coordinate unit of RoPE provides a more reliable spatial prior than directly using raw token-grid offsets. Table~\ref{tab:ablation_geocalibration} further shows that naive physical scaling is insufficient: directly multiplying position indices in RoPE with GSD  (i.e., $\gamma=0$ with GCC) reduces the average score from 73.78 to 73.70. This suggests that overly aggressive physical rescaling can destabilize the rotary response. 
A logarithmic calibration alternative~\cite{liu2022swin} improves the score to 74.24 by compressing the scale range, but still falls behind the full GeoRoPE setting with the proposed GCC. 
These results show that GCC is not merely injecting GSD information; rather, it provides a tempered coordinate transformation that balances ground-distance consistency with the stability of the pretrained prior.

\noindent\textbf{Effectiveness of GFC Module.} 
With the injection branch and GCC enabled, introducing GFC improves the average performance from 73.93 to 74.96, showing that a fixed RoPE frequency response is insufficient even after coordinate calibration. 
Table~\ref{tab:ablation_frequency_modulation} further compares GFC with a static learnable frequency scaling baseline, where a frequency-wise scale vector is optimized during fine-tuning but shared across all inputs and token pairs.
Although this slightly improves the average score from 73.93 to 73.95, it remains far below GeoRoPE with GFC. 
This indicates that the gain does not simply come from adding extra frequency parameters, but from relation-specific calibration that adapts positional sensitivity to scene-dependent spatial granularity. 


\noindent\textbf{Sensitivity to the Calibration Exponent $\gamma$.}
We analyze the effect of the GCC tempering exponent $\gamma$ in Fig.~\ref{fig:gamma}. 
Smaller $\gamma$ enforces stronger ground-distance calibration, while $\gamma=1$ approaches the original grid-relative RoPE coordinate system. 
GeoRoPE achieves the best overall performance at $\gamma=0.5$, reaching 84.21 Avg.C and 65.71 Avg.S. 
Classification remains relatively stable over a broad range of $\gamma$ between 0.5 and 0.75, suggesting that global semantic recognition is less sensitive to moderate changes in coordinate calibration. 
In contrast, segmentation exhibits a sharper optimum at $\gamma=0.5$: overly strong calibration may amplify rotary phases, whereas weak calibration leaves token-grid offsets insufficiently aligned with ground distance. 
These results support the use of a moderate $\gamma$, which balances ground-distance consistency with the stability of the pretrained prior.

\begin{wrapfigure}{r}{0.45\textwidth} 
    \centering
    \vspace{-15pt} 
    \includegraphics[width=\linewidth]{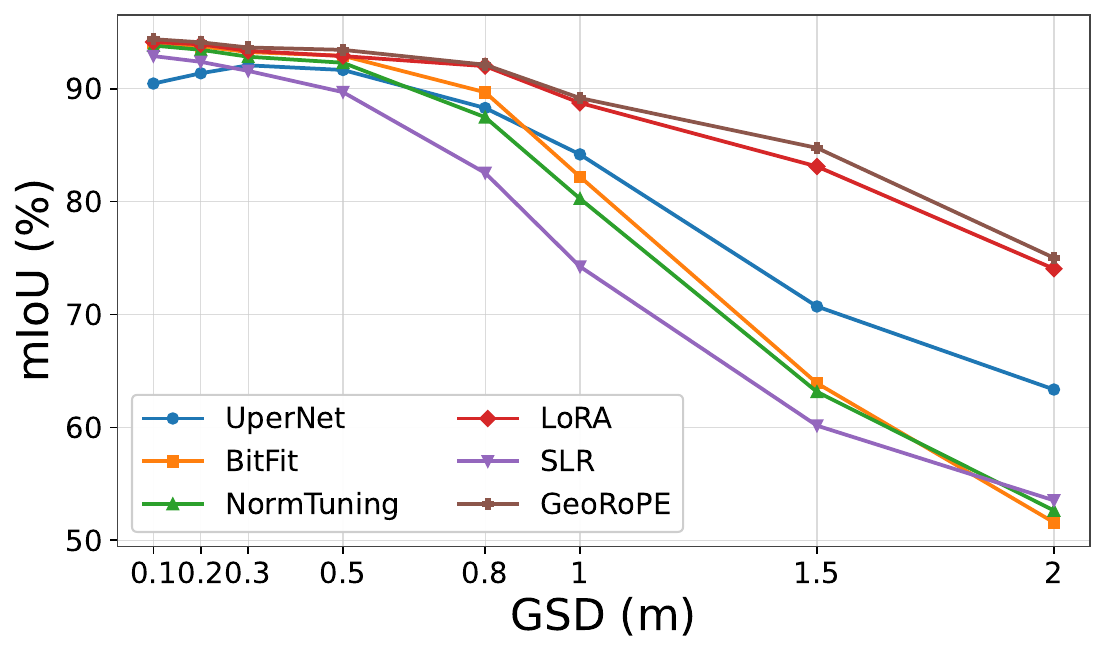}
    \vspace{-15pt} 
    \caption{Performance of adaptation methods across varying GSDs on segmentation.}
    \label{fig:gsd_comparison}
    \vspace{-10pt} 
\end{wrapfigure}


\noindent\textbf{Robustness to GSD Variation.}
We evaluate robustness under GSD shifts by resampling images from the PV4GER dataset to different effective GSDs for segmentation. 
As shown in Fig.~\ref{fig:gsd_comparison}, all methods degrade as the effective GSD becomes coarser, reflecting the loss of fine-grained spatial details for segmentation. 
Nevertheless, GeoRoPE consistently achieves the best performance and exhibits a gentler degradation trend than competing PEFT methods. 
This indicates that GeoRoPE help stabilize spatial predictions under GSD-induced resolution shifts. More results on classification are provided in Appendix.~\ref{appendix:GSD}.

\section{Conclusion}

We presented {GeoRoPE}, a RoPE-compatible and parameter-efficient spatial adaptation method for RSFMs. 
GeoRoPE addresses the scale ambiguity of token-grid positional interactions by recalibrating rotary positions from two complementary perspectives. 
Geo-Coordinate Calibration grounds relative offsets according to the ground distance represented by one token-grid step, while Geo-Frequency Calibration adjusts positional sensitivity to scene-dependent spatial granularity. 
Implemented through a lightweight attention-parallel adapter, GeoRoPE preserves the pretrained spatial prior of RSFMs while injecting geo-aware positional corrections. 
Extensive experiments demonstrate its effectiveness for cross-resolution remote-sensing adaptation.


\bibliographystyle{unsrtnat}
\bibliography{references}

\newpage
\addtocontents{toc}{\protect\setcounter{tocdepth}{2}}
\appendix
\section*{Appendix}

\tableofcontents
\newpage

\section{Broader Impact}
\label{appendix:broader_impact}

Remote sensing foundation models (RSFMs) are increasingly used for Earth observation tasks such as land-cover mapping, disaster monitoring, urban analysis, agricultural assessment, and environmental change detection. By improving the adaptation of RSFMs under heterogeneous ground sample distances (GSDs), GeoRoPE has the potential to make RSFMs reliable across sensors, regions, and downstream task settings. In particular, scale-aware positional adaptation can benefit applications where physically meaningful spatial relations are critical, including small-object mapping in high-resolution imagery, crop-field segmentation, infrastructure monitoring, and flood or wildfire damage assessment.

The proposed framework is designed as a lightweight positional adaptation mechanism rather than a new large-scale pretraining pipeline. Therefore, it may reduce the need to repeatedly pretrain or fully fine-tune large RSFMs for each new sensor or resolution regime. This can lower computational cost and energy consumption, especially when adapting existing backbones to new datasets. Such parameter-efficient adaptation is particularly relevant for research groups or public-sector organizations with limited computational resources.

\section{Limitation}
Despite its effectiveness, GeoRoPE still has several limitations. First, its physical calibration mainly relies on the availability of sensor metadata, while more complex imaging factors such as terrain relief, or temporal changes are not addressed. Second, the semantic-guided frequency modulation is learned from downstream data and may be sensitive to dataset-specific spatial textures when the target data are limited. Future work will extend GeoRoPE toward uncertainty-aware geometric calibration and broader multi-sensor, multi-temporal remote sensing scenarios.

\section{Method Details}
\label{appendix:method_details}


Standard RoPE relies on discrete grid indices, formulating the positional phase as $\phi_{\boldsymbol{\Delta}_{\mathbf u,\mathbf v}, m} = \boldsymbol{\Delta}_{\mathbf u,\mathbf v} \cdot \theta_m$, which inherently ignores the physical footprint of spatial tokens. 
In this standard encoding, the angular frequency $\theta_m$ inherently corresponds to a continuous spatial wavelength $L_m = \frac{2\pi}{\theta_m}$. This wavelength dictates the positional sensitivity and scale characteristics of the model: shorter wavelengths capture fine-grained local shifts, while longer wavelengths encode coarse, global spatial relationships.

To inject physical scale awareness into the model, we anchor the effective wavelength to the real-world spatial resolution. Let $\ell^{(l)}$ denote the physical ground distance per token step at stage $l$. To ensure the model accurately captures the finest positional variations, the target wavelength $p_s$ is bounded by the physical resolution:

\begin{equation}
    p_s = \frac{2\pi}{\rho} \cdot \ell^{(l)}
\end{equation}
where $\rho$ is the target phase difference (typically $\rho=1$ rad) for adjacent tokens at the finest scale. To guarantee that adjacent token features remain distinguishable under the finest resolution, we define the baseline reference physical spacing as $\ell_{\mathrm{ref}}^{(l)}$.

For a heterogeneous sensor with a coarser ground distance $\ell^{(l)}$, the physical distance per token step exceeds $\ell_{\mathrm{ref}}^{(l)}$. Therefore, instead of scaling the predefined frequency bank $\theta_m$, we apply Geo-Coordinate Calibration to project the discrete relative grid offset $\boldsymbol{\Delta}_{\mathbf u,\mathbf v}$ into a unified geo-calibrated relative coordinate $\widetilde{\boldsymbol{\Delta}}_{\mathbf u,\mathbf v}$:

\begin{equation}
    \widetilde{\boldsymbol{\Delta}}_{\mathbf u,\mathbf v} = \rho_G^{(l)} \boldsymbol{\Delta}_{\mathbf u,\mathbf v}, \quad \text{where} \quad \rho_G^{(l)} = \left(\frac{\ell^{(l)}}{\ell_{\mathrm{ref}}^{(l)}}\right)^{1-\gamma}
\end{equation}

While the baseline reference physical spacing $\ell_{\mathrm{ref}}^{(l)}$ guarantees that positional variations are perfectly resolvable at the finest available resolution, we must also ensure spatial stability when the model encounters significantly coarser resolutions. Specifically, the maximum phase rotation at the highest frequency dimension must strictly not exceed $\pi$ to prevent severe phase aliasing (the Nyquist limit).

To enforce this constraint across the entire resolution spectrum, we establish the following inequality for the maximum phase shift $\Delta \phi_{\max}^{(l)}$ under the maximum spatial footprint $\ell_{\max}^{(l)}$:

\begin{equation}
    \Delta \phi_{\max}^{(l)} = 1 \cdot \left( \frac{\ell_{\max}^{(l)}}{\ell_{\mathrm{ref}}^{(l)}} \right)^{1-\gamma} \cdot \theta_0 \le \pi
\end{equation}

Given that the highest angular frequency in the standard basis is $\theta_0 = 1$, we can simplify the expression. By taking the natural logarithm of both sides, we obtain:

\begin{equation}
    (1 - \gamma) \ln\left( \frac{\ell_{\max}^{(l)}}{\ell_{\mathrm{ref}}^{(l)}} \right) \le \ln(\pi)
\end{equation}

By rearranging the terms, we derive the explicit theoretical lower bound for the scale control factor $\gamma$:

\begin{equation}
    \label{eq:gamma_lower_bound} \gamma \ge 1 - \frac{\ln(\pi)}{\ln(\ell_{\max}^{(l)} / \ell_{\mathrm{ref}}^{(l)})}
\end{equation}

\textbf{Analysis of the Bound:}
When the resolution disparity is minimal ($\ell_{\max}^{(l)}/\ell_{\mathrm{ref}}^{(l)} \le \pi$), the model does not require explicit scale compression to satisfy the Nyquist limit, and the condition $\gamma \ge 0$ naturally holds.
When the resolution gap is massive (e.g., a $100\times$ difference), the formula calculates a definitive theoretical lower bound.
Ultimately, as long as $\gamma$ is set above this mathematically derived threshold, the model absolutely safely circumvents the risk of high-frequency aliasing while maximally preserving the physical scale variance across heterogeneous remote sensing representations.

\begin{table*}[t]
    \centering
    \caption{
    \textbf{Summary of evaluation datasets.}
    We use the full GEO-Bench benchmark as the main evaluation suite, including six classification datasets and six semantic segmentation datasets. S1 denotes Sentinel-1, S2 denotes Sentinel-2, and NIR denotes near-infrared.
    }
    \label{tab:dataset_summary}
    \scriptsize
    \setlength{\tabcolsep}{3.2pt}
    \renewcommand{\arraystretch}{1.12}
    \begin{tabular}{
        >{\raggedright\arraybackslash}p{0.15\textwidth}
        >{\raggedright\arraybackslash}p{0.13\textwidth}
        >{\raggedright\arraybackslash}p{0.22\textwidth}
        >{\raggedright\arraybackslash}p{0.18\textwidth}
        >{\raggedright\arraybackslash}p{0.09\textwidth}
        >{\raggedright\arraybackslash}p{0.13\textwidth}
    }
        \toprule
        \textbf{Dataset} 
        & \textbf{Task Type} 
        & \textbf{Task} 
        & \textbf{Sensor / Modality} 
        & \textbf{GSD} 
        & \textbf{Image Resolution / Tile Size} \\
        \midrule
        \textbf{m-bigearthnet} 
        & Classification 
        & Multi-label land-cover classification 
        & Sentinel-2 
        & 10 m 
        & 120 $\times$ 120  \\
        
        \textbf{m-so2sat} 
        & Classification 
        & Local climate zone classification 
        & Sentinel-1 + Sentinel-2 
        & 10 m 
        & 32 $\times$ 32  \\
        
        \textbf{m-brick-kiln} 
        & Classification 
        & Brick kiln binary classification 
        & Sentinel-2 
        & 10 m 
        & 64 $\times$ 64  \\
        
        \textbf{m-forestnet} 
        & Classification 
        & Deforestation driver classification 
        & Landsat-8 
        & 15 m 
        & 332 $\times$ 332  \\
        
        \textbf{m-eurosat} 
        & Classification 
        & Land-use / land-cover classification 
        & Sentinel-2 
        & 10 m 
        & 64 $\times$ 64  \\
        
        \textbf{m-pv4ger} 
        & Classification 
        & Photovoltaic / solar panel binary classification 
        & RGB aerial imagery 
        & 0.1 m 
        & 320 $\times$ 320  \\
        
        \midrule
        
        \textbf{m-pv4ger-seg} 
        & Semantic segmentation 
        & Photovoltaic / solar panel segmentation 
        & RGB aerial imagery 
        & 0.1 m 
        & 320 $\times$ 320  \\
        
        \textbf{m-chesapeake-landcover} 
        & Semantic segmentation 
        & Land-cover segmentation 
        & RGB + NIR 
        & 1 m 
        & 256 $\times$ 256  \\
        
        \textbf{m-cashew-plantation} 
        & Semantic segmentation 
        & Cashew plantation segmentation 
        & Sentinel-2 
        & 10 m 
        & 256 $\times$ 256  \\
        
        \textbf{m-SA-crop-type} 
        & Semantic segmentation 
        & Crop-type segmentation 
        & Sentinel-2 
        & 10 m 
        & 256 $\times$ 256  \\
        
        \textbf{m-nz-cattle} 
        & Semantic segmentation 
        & Cattle segmentation 
        & RGB aerial imagery 
        & 0.1 m 
        & 500 $\times$ 500  \\
        
        \textbf{m-NeonTree} 
        & Semantic segmentation 
        & Tree crown segmentation 
        & RGB + hyperspectral + elevation 
        & 0.1 m 
        & 400 $\times$ 400  \\
        \bottomrule
    \end{tabular}
\end{table*}

\section{Experimental Details}
\label{appendix:experimental_details}

This section describes the experimental protocol, evaluation metrics, datasets, backbones, and implementation details used in our experiments.

\subsection{Dataset Details}
\label{appendix:dataset_details}

We evaluated GeoRoPE on a diverse collection of remote sensing datasets covering classification and segmentation tasks. The classification datasets include BigEarthNet, Brick Kiln, EuroSAT, ForestNet, PV4GER, and So2Sat. The segmentation datasets include Cashew Plantation, Chesapeake Land Cover, NeonTree, NZ Cattle, PV4GER-Seg, and SA Crop. These datasets cover different sensors, spatial resolutions, land-cover types, and geographic regions, making them suitable for evaluating scale-aware and region-heterogeneity-aware adaptation. 
For multi-resolution datasets, Table~\ref{tab:dataset_summary} reports the spatial resolution used in the GEO-Bench evaluation setting.

\subsection{Backbone Details}
\label{appendix:backbone_details}

We evaluated GeoRoPE on multiple representative RSFMs with different positional encoding designs. 
As summarized in Table~\ref{appendix:backbone_details}, for backbones pretrained on heterogeneous remote sensing sources, the reported GSD range denotes the nominal spatial-resolution coverage of the corresponding pretraining sensors or datasets. 
For datasets with image-level metadata, such as fMoW-RGB, the GSD varies across samples. 
For multimodal models such as DOFA and Satlas, the range summarizes the spatial resolutions of their supported pretraining modalities rather than a single fixed input resolution.

\subsection{Training Details}
\label{appendix:training_details}
All models are optimized using the AdamW optimizer. However, the learning rate (LR) scheduling strategies vary across different backbones to accommodate their specific pretraining characteristics. For the DOFA backbone, we employ a \texttt{CosineAnnealing} LR schedule. In contrast, for ScaleMAE and SatLas, we utilize a \texttt{ReduceLROnPlateau} scheduler, which adjusts the learning rate based on the validation performance.
The training batch sizes are standardized according to the downstream tasks. For classification tasks, we use a batch size of 32 across all experiments. For semantic segmentation tasks, the batch size is set to 16 to balance the memory constraints associated with high-resolution feature maps.

During adaptation, considering that ScaleMAE is originally pretrained on standard RGB imagery, specific adaptation is required for multispectral data. To maintain consistency with the input dimensionality of the pretrained ScaleMAE weights, we selectively extract the corresponding spectral channels from the multispectral input to align with the RGB channels used during its pretraining phase. Across all backbones, GeoRoPE is applied at the attention side without disrupting the original pretrained positional priors. It is noted that, 
for models like SatLas, which are based on the Swin-Transformer architecture, the attention mechanism is inherently restricted to local windows. To ensure compatibility, we explicitly construct a local spatial grid independently within each attention window. The GeoRoPE modulation is consequently applied using these window-local relative positions, allowing us to maintain robust geographic grounding while adhering strictly to the localized computation constraints of the Swin-Transformer.

\begin{table*}[t]
    \centering
    \caption{
    \textbf{Summary of evaluated remote sensing foundation model backbones.}
    The selected backbones cover different pretraining datasets, sensor modalities, spatial-resolution ranges, and positional encoding designs.
    }
    \label{tab:appendix_backbone_details}
    \footnotesize
    \setlength{\tabcolsep}{3.5pt}
    \renewcommand{\arraystretch}{1.12}
    \begin{tabular}{
        >{\raggedright\arraybackslash}p{0.12\textwidth}
        >{\raggedright\arraybackslash}p{0.23\textwidth}
        >{\raggedright\arraybackslash}p{0.20\textwidth}
        >{\raggedright\arraybackslash}p{0.23\textwidth}
        >{\raggedright\arraybackslash}p{0.13\textwidth}
    }
        \toprule
        \textbf{Backbone} 
        & \textbf{Pretraining Dataset(s)} 
        & \textbf{Input Modality} 
        & \textbf{GSD / Spatial Resolution Range} 
        & \textbf{Original PE} \\
        \midrule
        
        \textbf{ScaleMAE} 
        & fMoW-RGB 
        & RGB optical imagery 
        & Approx. 0.3--1.5 m; image-level GSD varies across samples 
        & GSD-aware PE \\
        
        \textbf{DOFA} 
        & Sentinel-1, Sentinel-2, Gaofen-2, NAIP, and EnMAP 
        & SAR, multispectral, RGB aerial, and hyperspectral imagery 
        & Approx. 1--30 m; covering high-resolution aerial imagery, medium-resolution optical/SAR, and hyperspectral imagery 
        & APE-style encoding \\
        
        \textbf{Satlas} 
        & SatlasPretrain, including Sentinel-2, Sentinel-1, Landsat 8/9, and aerial imagery 
        & Optical, SAR, and aerial imagery depending on the released checkpoint 
        & Approx. 0.5--30 m; 0.5--2 m aerial imagery, 10 m Sentinel imagery, and 30 m Landsat imagery 
        & Relative PE \\
        
        \bottomrule
    \end{tabular}
\end{table*}

\subsection{Compute Resources}
\label{appendix:compute_resources}

All training and evaluation experiments are conducted on three NVIDIA A6000 GPUs, each with 48GB of memory, using a batch size of 16. The training machine is equipped with an Intel(R) Xeon(R) Gold 6336Y CPU @ 2.40GHz. 

\section{Supplementary Results}
\label{appendix:supp_results}


\begin{table*}[t]
    \centering
    \caption{
    \textbf{Ablation study of GeoRoPE.}
    Avg.C, Avg.S, and Avg.P denote the macro-average performance over classification datasets, segmentation datasets, and all datasets, respectively.
    Bold indicates the best performance.
    }
    \label{tab:ablation_georope}

    \resizebox{\textwidth}{!}{%
    \begin{tabular}{l cccccc cccccc ccc}
        \toprule
        \multirow{2}{*}{\textbf{Variant}}
        & \multicolumn{6}{c}{\textbf{Classification}}
        & \multicolumn{6}{c}{\textbf{Segmentation}}
        & \multicolumn{3}{c}{\textbf{Average}} \\
        \cmidrule(lr){2-7} \cmidrule(lr){8-13} \cmidrule(lr){14-16}
        & BE & BK & ES & FN & PV & SS
        & CW & NT & CT & CP & PS & SC
        & Avg.C & Avg.S & Avg.P \\
        \midrule

        LoRA
        & 88.13 & 98.50 & 96.92 & \textbf{54.24} & 97.99 & 60.53
        & \textbf{68.61} & 55.97 & 75.26 & 60.56 & 94.14 & 33.57
        & 82.72 & 64.69 & 73.70 \\

        RoPE
        & 88.10 & \textbf{98.59} & 97.37 & 50.74 & 98.29 & 62.37
        & 67.17 & 55.47 & 75.18 & \textbf{61.88} & 94.23 & 34.40
        & 82.58 & 64.72 & 73.65 \\

        w/ GCC
        & 90.14 & 98.30 & 96.93 & 52.55 & \textbf{98.30} & 61.66
        & 66.46 & 58.93 & 75.35 & 59.91 & 94.31 & 34.27
        & 82.98 & 64.87 & 73.93 \\

        w/ GFC
        & 88.54 & 98.30 & \textbf{97.44} & 51.50 & 98.09 & 62.07
        & 66.98 & 56.78 & 75.24 & 61.21 & 94.34 & 34.79
        & 82.66 & 64.89 & 73.77 \\

        \rowcolor{gray!10}
        \textbf{GeoRoPE (full)}
        & \textbf{94.16} & 98.48 & \textbf{97.44} & 54.21 & 98.21 & \textbf{62.78}
        & 68.48 & \textbf{59.19} & \textbf{75.45} & 61.54 & \textbf{94.39} & \textbf{35.19}
        & \textbf{84.21} & \textbf{65.71} & \textbf{74.96} \\

        \bottomrule
    \end{tabular}%
    }
\end{table*}
\subsection{Detailed Ablation}

Here we provide a more complete ablation analysis across all datasets.
As shown in Table~\ref{tab:ablation_georope}, adding a vanilla RoPE branch over LoRA brings only marginal improvements and even degrades several datasets such as ForestNet, Cashew, and NeonTree, showing that index-based relative positions alone cannot reliably resolve physical scale shifts and local spatial heterogeneity.
Building upon RoPE, GCC improves Avg.P from 73.65 to 73.93, with clear gains on BigEarthNet and NeonTree, indicating its effectiveness in calibrating token relations under varying physical spatial coverages.
GFC further benefits structure-sensitive segmentation datasets such as pv4ger-seg and SA\_Crop, suggesting that semantic-guided frequency modulation helps adapt positional responses to local spatial complexity.
The full GeoRoPE achieves the best Avg.C, Avg.S, and Avg.P, demonstrating that global geo-calibration and adaptive semantic modulation are complementary: GCC enforces physical scale consistency, while GFC adjusts the effective positional frequency for heterogeneous regions.


\begin{figure}[t]
    \centering 
    \includegraphics[width=\linewidth]{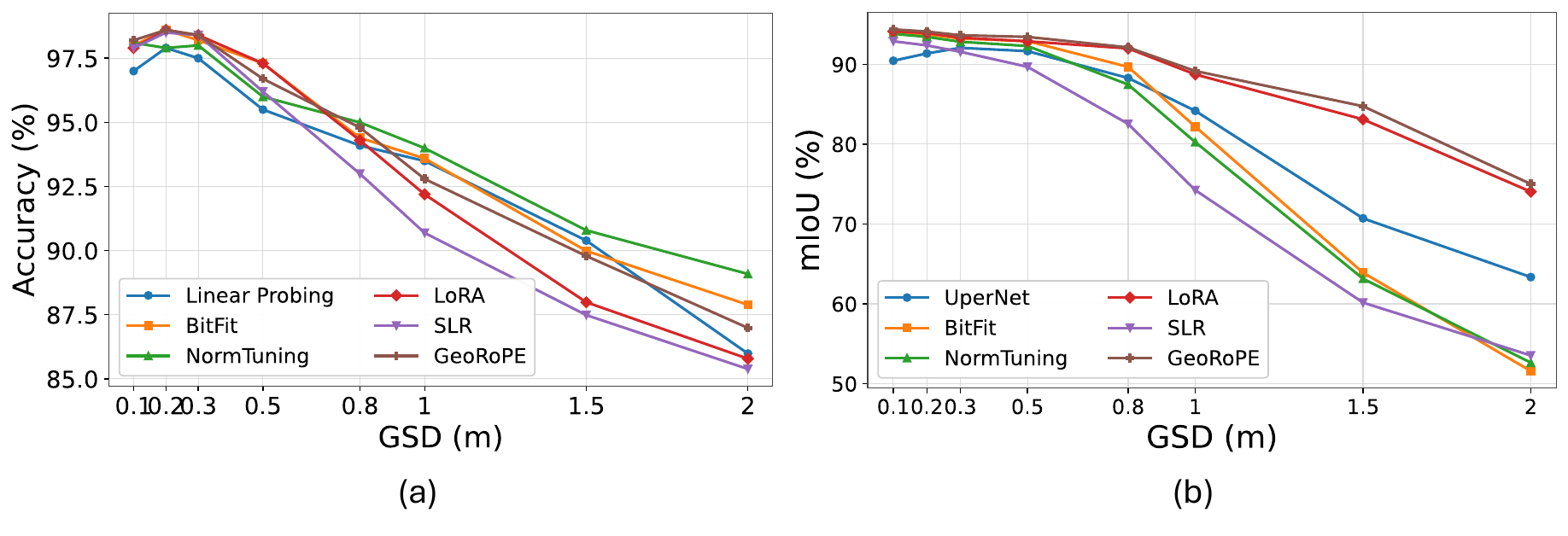}
        \caption{\textbf{Performance curves under varying GSDs.} (a) classification accuracy on pv4ger. (b) segmentation mIoU on pv4ger\_seg. 
        }   
\label{fig:gsd_comparison_all}
\end{figure}

\subsection{Robustness to GSD variations.}
\label{appendix:GSD}
We further evaluate the robustness of different adaptation methods under varying simulated GSDs on two PV4Ger benchmarks, including classification on \textit{pv4ger} and segmentation on \textit{pv4ger\_seg}. Both datasets use images with a native GSD of 0.1\,m and an input resolution of $320\times320$. 
As shown in Fig.~\ref{fig:gsd_comparison_all}, classification is overall more robust to GSD changes than segmentation. In particular, when the GSD is slightly increased to 0.2\,m, several methods, including LoRA, SLR, and GeoRoPE, even show a small performance gain, suggesting that mild resampling may suppress noise while preserving sufficient semantic cues for classification. 

For classification, most methods remain relatively stable at small-to-moderate GSDs, but their performance diverges more clearly as the GSD becomes larger. Although LoRA and BitFit achieve strong accuracy at small GSDs, they degrade more rapidly at coarse resolutions. In contrast, GeoRoPE-LoRA shows stronger robustness under large GSD shifts and achieves the best accuracy in the high-GSD regime, indicating better preservation of scale-invariant semantic representations. 
For segmentation, all methods suffer more severe degradation due to the loss of fine-grained boundary cues, while GeoRoPE consistently maintains the best robustness.

\begin{figure}[t]
    \centering 
    \includegraphics[width=\linewidth]{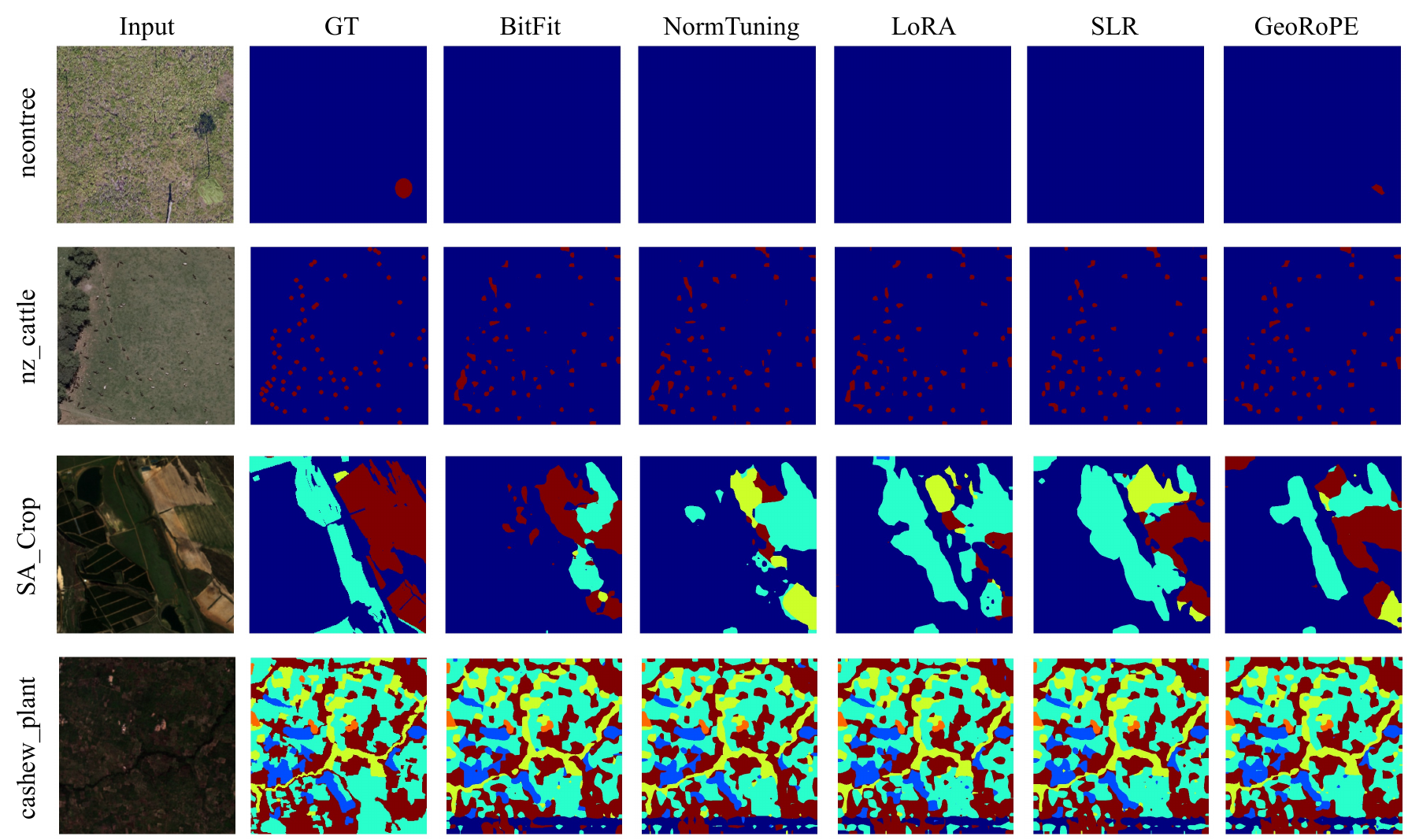}
        \caption{Qualitative comparison of different PEFT methods on representative segmentation datasets.
        }   
\label{fig:segmentation_2}
\end{figure}

\subsection{More Visualizations}
\label{appendix:visualizaitons}

We provide qualitative segmentation comparisons in Fig.~\ref{fig:segmentation_2}. For tiny, sparse targets in heterogeneous backgrounds (i.e., NeonTree and NZ Cattle), most PEFT baselines (e.g., BitFit, NormTuning) produce empty or spatially ambiguous predictions. Conversely, GeoRoPE successfully localizes these objects, demonstrating that adaptive frequency modulation preserves fine-grained sensitivity without introducing excessive noise. 
For region-level datasets (i.e., SA Crop and Cashew Plant), baselines often generate fragmented parcels and confused boundaries. GeoRoPE instead yields spatially coherent masks and sharper semantic boundaries. Overall, GeoRoPE effectively handles both tiny object localization and complex region delineation.


\end{document}